\documentclass[runningheads]{llncs}
\usepackage{graphicx}
\usepackage{subfig}

\begin{document}
\title{Checkworthiness in Automatic Claim Detection Models: Definitions and Analysis of Datasets}
\titlerunning{Checkworthiness in Automatic Claim Detection Models}
%
\author{Liesbeth Allein*\orcidID{0000-0002-7776-2156} \and
Marie-Francine Moens\orcidID{0000-0002-3732-9323}}
\authorrunning{L. Allein and M. Moens}
%
\institute{KU Leuven, Leuven, Belgium \\
\email{liesbeth.allein@kuleuven.be}; \email{sien.moens@kuleuven.be}}
\maketitle              
\begin{abstract}
Public, professional and academic interest in automated fact-checking has drastically increased over the past decade, with many aiming to automate one of the first steps in a fact-check procedure: the selection of so-called checkworthy claims. However, there is little agreement on the definition and characteristics of checkworthiness among fact-checkers, which is consequently reflected in the datasets used for training and testing checkworthy claim detection models. After elaborate analysis of checkworthy claim selection procedures in fact-check organisations and analysis of state-of-the-art claim detection datasets, checkworthiness is defined as the concept of having a spatiotemporal and context-dependent worth and need to have the correctness of the objectivity it conveys verified. This is irrespective of the claim's perceived veracity judgement by an individual based on prior knowledge and beliefs. Concerning the characteristics of current datasets, it is argued that the data is not only highly imbalanced and noisy, but also too limited in scope and language. Furthermore, we believe that the subjective concept of checkworthiness might not be a suitable filter for claim detection. 

\end{abstract}
\section{Introduction}


Fact-checking coverage has been diffusing rapidly in U.S. political media over the past few years, with media emphasizing a journalist's supposed professional role and status as 'truth-seeker' to the public \cite{graves:16}. Not only has fact-checking seen an increase in news media, but also the number and spread of independent fact-check organisations and organisations linked to news papers around the world has drastically risen and expanded over the past few years. The number has nearly quintupled from 44 active fact-checkers primarily in the U.S. and Europe in 2014 \cite{reporters:14} to around 210 active fact-check organisations in 68 countries all over the world in 2019 \cite{reporters:19}. Research, more specifically computational journalism and computer science, has joined this growing trend and has been exploring the automation of fact-check processes. Several international workshops have been organised to tackle computational fact-check models. In the Workshop on Fact Extraction and VERification (FEVER), for example, participants are challenged to create classifiers that can predict whether information retrieved from Wikipedia pages supports or refutes human-written factoid claims \cite{thorne:18b}. For the CheckThat! Lab at the Conference and Labs of the Evaluation Forum (CLEF), several research groups have built verification models that rank given web pages based on their usefulness for fact-checking a given claim, classify those pages, identify useful passages in the texts and, eventually, predict the claim's factuality based on those retrieved passages \cite{hasanain:19}. In order to automatically decide which information needs to be fact-checked, researchers have been attempting to develop computational models that extract so-called \textit{checkworthy} claims from given texts. The checkworthiness of claims indicates that checkworthy claims are somehow distinguishable from other, non-checkworthy claims and that a claim's veracity can and should be recovered based on evidence extracted from other sources. However, the concept of \textit{checkworthiness} lacks a generally accepted and shared definition among fact-check organisations with many of them applying their own delineation to their fact-checking pipeline. The disagreement on the definition of checkworthiness and the difference between checkworthy and non-checkworthy claims, subsequently, permeates the datasets which are used to train and test checkworthy claim detection models, resulting in inconsistent and less reliable datasets. In this paper, we discuss the definition of checkworthiness in journalism, fact-check organisations and, more elaborately, computational claim detection models for fact-checking. Furthermore, we delve into the characteristics of several datasets and how these influence model performance. 

\section{Checkworthiness}


In this section, we aim at providing a more elaborate characterization of checkworthiness by looking at how it is defined in dictionaries, journalism, fact-check organizations and, consequently, claim detection models. Furthermore, we discuss the shortcomings of these characteristics in several datasets.
 
\subsection{General Definition}
Checkworthiness is not explicitly mentioned or defined in any of the large English dictionaries (Cambridge, Oxford, Collins, Merriam-Webster and Macmillan). Therefore, we split the concept and attempt to define it using the definitions of \textit{check}, \textit{worthiness} and \textit{worthy}. By checking a claim, a person - or in this case, a system - quickly verifies its correctness \cite{cambridge:check}. Worthiness can be defined as the quality of deserving to be treated in a specified manner or deserving attention, on the one hand, and as the worth, value and suitability of something, on the other \cite{oxford:worthiness,cambridge:worthiness}. A checkworthy claim thus deserves to be treated in a specified manner - in this case, it should be checked - and attention should be paid to it. Moreover, a worthy claim has a certain worth, value and suitability which is relative in respect to its importance, usefulness and merit \cite{oxford:worth}. In all, checkworthiness can be defined as the characteristic of entailing a certain worth, value, suitability and, especially, need to have the correctness of what it conveys verified. That definition of checkworthiness, however, is still rather abstract. Due to a lacking generally accepted definition of checkworthiness, researchers resort to their own interpretation and delineation and/or on those specified by journalists and fact-checkers. 
\subsection{Datasets and Computational Models}

The divergent interpretations of checkworthiness are reflected in the datasets used for training and testing checkworthy claim detection models. In this section, we discuss the characteristics of the ClaimBuster \cite{hassan:17,hassan:15}, CT-CWC-18 \cite{nakov:18}, CT19-T1 \cite{atanasova:19}, CW-USPD-2016 \cite{gencheva:17} and TATHYA \cite{patwari:17} datasets and how they define checkworthiness. Furthermore, we look at how these datasets are constructed and briefly mention some models that use these datasets. These datasets will be further analyzed in the remainder of this paper.

\subsubsection{ClaimBuster}
The team behind ClaimBuster \cite{hassan:17,hassan:15} constructed a dataset of non-factual (NFS), unimportant factual (UFS) and checkworthy factual sentences (CFS) taken from American presidential debate transcripts \cite{hassan:15}. The NFS category contains subjective sentences such as opinions, beliefs and questions. As for the UFS and CFS, they argue that the difference between the two is a difference in checkworthiness, with checkworthiness being rather generally defined as the present appeal of the general public to recover a claim's truthfulness. The sentences were randomly labeled by paid journalists, professors and university students with basic knowledge of U.S. politics, each of them requiring to label a number of randomly assigned screening sentences to check their labeling quality. A score between 0 and 1 is assigned to each participant based on the label agreement between them and three domain experts. Participants with high labeling agreement were named top-quality participants. In total, presidential debate transcripts from 2016, 2012, 2008 and 2004 were labeled, with each sentence labeled by at least two participants. Only sentences with an agreed label by two top-quality participants were used in the training and evaluation set, resulting in a dataset of 1,571 sentences (882 NFS, 252 UFS, 437 CFS) in 2015 \cite{hassan:15} and 20,617 sentences (13,671 NFS, 2097 UFS, 4849 CFS) in 2017 \cite{hassan:17}. 

Feature-based models such as Support Vector Machine (SVM), Naive Bayes Classifier and Random Forest Classifier were implemented for the classification task, with the models trained on the larger dataset \cite{hassan:17} outperforming those trained on the smaller dataset \cite{hassan:15}. SVM obtained the highest results for both datasets. Jiminez and Li \cite{jiminez:18} took a neural approach and computed a CNN + BiLSTM model and trained the model on a hand-labeled dataset of 8,231 sentences with NFS, UFS and CFS labels. They looked at the difference in model performance when the model needs to predict three classes and two classes (where NFS and UFS are concatenated in one class). It appears that, in this case, model performance is higher for all three classes when the claim detection task is approached as a three-class classification problem. 

\begin{table}[]
    \centering
    \begin{tabular}{|c|c|c|c|c|}
        \hline
        Model & Precision & Recall & F1-score & Dataset Size: \\
         & CFS & CFS & CFS & Number of sentences \\
        \hline
        \hline
        SVM \cite{hassan:15} & 0.69 & 0.65 & 0.67 & 1571 \\
        \hline
        SVM \cite{hassan:17} & 0.72 & 0.67 & 0.70 & 20,617 \\
        \hline
        CNN + BiLSTM \cite{jiminez:18} & & & & 8231 \\
        Three classes (NFS, UFS, CFS) & 0.68 & 0.65 & 0.61 & \\
        Two classes (NFS + UFS, CFS) & 0.60 & 0.59 & 0.59 & \\
        \hline
    \end{tabular}
    \caption{Overview of model performance using the ClaimBuster dataset}
    \label{tab:claimbuster}
\end{table}

\subsubsection{CT-CWC-18}
For the first CLEF CheckThat! Lab in 2018, the CT-CWC-18 was constructed and contains U.S. presidential debate transcripts and, additionally, Donald Trump speeches \cite{atanasova:18,nakov:18}. Instead of manually labeling non-factual, unimportant factual and checkworthy factual sentences, they approached it as a binary classification task and automatically assigned two labels - checkworthy and non-checkworthy - to the dataset. The gold standard labels were automatically derived from the analysis carried out by FactCheck.org. If a claim was fact-checked, it was labeled as checkworthy. However, the FactCheck annotations did not consistently cover whole sentences or sometimes exceeded sentence boundaries. In order to have sentence-level annotations, the authors gave the label of an annotated sentence part to the entire sentence and to all the sentences containing a part of the annotation. In total, the dataset contains 8,946 sentences of which 282 are labeled checkworthy. They also created an equivalent Arabic dataset by translating the English dataset. 

Two models that outperformed both baselines (random permutation of the input sentences and n-gram based classifier) are a multilayer perceptron with feature-rich representation \cite{zuo:18} and a recurrent neural network \cite{hansen:18}. Zuo, Karakas and Banerjee~\cite{zuo:18} took a hybrid approach and combined simple heuristics for assigning a checkworthiness score to each sentence with SVMs, multilayer perceptrons and an ensemble model combining the two supervised machine learning techniques. The claim detection model of Hansen et al.~\cite{hansen:18} transforms the input claims to sentence embeddings learnt by a Recurrent Neural Network, which are subsequently sent to a Recurrent Neural Network with GRU memory units. The official evaluation measures were, among others, Mean Average Precision (MAP), Mean Reciprocal Rank (MRR) and Mean R-Precision (MR-P). Results of the two models are given in Table \ref{tab:clef}.

\begin{table}[]
    \centering
    \begin{tabular}{|c|c|c|c|c|}
        \hline
        Model & MAP & MRR & MR-P & Dataset \\
        \hline
        \hline
        Multilayer perceptron \cite{zuo:18} & 0.1332 & 0.4965 & 0.1352 & CT-CWC-18 \\
        RNN + GRU \cite{hansen:18} & 0.1152 & 0.3159 & 0.1100 & CT-CWC-18 \\
        \hline
        \hline
        LSTM \cite{hansen:19} & 0.1660 & 0.4176 & 0.1387 & CT19-T1 \\
        Feedforward NN \cite{favano:19} & 0.1597 & 0.1953 & 0.2052 & CT19-T1 \\
        \hline
    \end{tabular}
    \caption{Overview of model performance in the CLEF CheckThat! Lab using CT-CWC-18 and CT19-T1 datasets}
    \label{tab:clef}
\end{table}

\subsubsection{CT19-T1}
For the 2019 edition of the CheckThat! Lab, the CT-CWC-18 was extended with three press-conferences, six public speeches, six debates and one post, all annotated by factcheck organisation FactCheck.org. The extended dataset was named CT19-T1 \cite{atanasova:19}. However, instead of determining whether or not a sentence is checkworthy, the claim detection models are expected to rank sentences in terms of checkworthiness, with a higher rank indicating a higher level of checkworthiness. The models thus output a score between 0 (certainly non-checkworthy) and 1 (certainly checkworthy). Checkworthy sentences in the train set are, again, automatically identified based on the transcript analyses from FactCheck.org.

Hansen et al. \cite{hansen:19} have built further upon their RNN model \cite{hansen:18} in the 2018 workshop. Sentences are transformed to dual sentence representations capturing both semantics (Word2Vec word embeddings) and syntactic sentence structure (syntactic dependencies for each word) which are then fed to a LSTM neural network with an attention layer on top. Favano, Carman and Lanzi \cite{favano:19} built a feedforward neural network that takes Universal Sentence Encoder embeddings as input. They, additionally, experimented with various training data modifications. The performance of these results are shown in Table \ref{tab:clef}.

\subsubsection{CW-USPD-2016}
For the CW-USPD-2016 dataset \cite{gencheva:17}, a similar dataset and labeling method as in CT-CWC-18 and CT19-T1 are used, but the number of annotated transcripts is limited to one vice-presidential and three presidential debates. Gold standard labels are retrieved from nine reputable fact-check sources instead of one, resulting in 5,415 labeled sentences.
Their feedforward neural network has a MAP of 0.427.

\subsubsection{TATHYA}
As all the datasets discussed above, the TATHYA dataset \cite{patwari:17} collected transcripts of multiple political debates: seven Republican primary debates, eight Democratic primary debates, three presidential debates and one vice-presidential debate. They also included Donald Trump's Presidential Announcement Speech.
The period in which these debates occurred is not specified, however we can infer that they took place during the 2016 presidential campaign. A statement is labeled checkworthy if it is fact-checked by at least one of eight fact-check websites (Washington Post, factcheck.org, Politifact, PBS, CNN, NY Times, Fox News, USA Today). The dataset consists of 15,735 sentences of which 967 are checkworthy. 
Their SVM model obtains 0.227 precision, 0.194 recall and 0.209 F1-score on the held-out test set of the presidential debates. Their clustering multi-classifier system has 0.188 precision, 0.248 recall and 0.214 F1-score on the same test set. 

\subsection{Fact-Check Organisations} \label{factcheckorgan}


The fact-check organisation that is regularly used as label reference in the abovementioned datasets is FactCheck.org \cite{factcheck:process}. When fact-checkers at FactCheck.org perform fact-checking, they look for statements that are based on facts. Once they suspect that a factual, objective statement could contain disinformation, misinformation, deceit or inaccuracy, they contact the person or organisation who uttered the statement and ask for clarification and evidence to support the claim. If the given evidence is insufficient or inaccurate, they conduct their fact-check procedure. As a result, the claims for which sufficient, truthful evidence is provided are not explicitly marked on their website; only claims that underwent the full fact-check procedure are indicated. Moreover, the checked topics and persons/organisations slightly depend on the election cycle, with a focus on presidential candidates, Senate candidates and members of Congress during presidential elections, midterm elections and off-election years, respectively. In all, it can be said that FactCheck.org selects checkworthy claims based on their objectivity, their suspected falsity and the election cycle at the time of fact-checking. Politifact, on the other hand, states on their website that they choose a statement to fact-check based on its verifiability, possibly misleading character, significance, transferability and truthfulness that could be questioned by common people. Furthermore, they claim to select statements about topics that are currently in the news, attempt to balance fact-check coverage on Democrats and Republicans, and focus on the political party currently in power and people who are prone at making attention-drawing and/or misleading statements \cite{politifact:18}. Contrary to choosing claims that are currently in the news, Snopes take a more reader-oriented stance and states that they write about topics that are in high demand or interest among their readers. They recover readers' interests by querying their search engine, reader submissions, comments and questions on their social media accounts and trending topics on Google and social media. They claim to avoid making personal judgements about a claim's importance, controversy, obviousness or depth \cite{snopes:transparency}. Full Fact states that it fact-checks claims about topics of national interest - such as economy, crime, health, immigration, education, law and Europe - for which reference sources exist and for which they have in-house expertise \cite{fullfact}. They do not discuss how they evaluate a claim's checkworthiness in more detail. Some fact-check organisations such as TruthorFiction \cite{truthorfiction} do not elaborate on how they choose the statements they fact-check, thus not providing any definition of checkworthiness.

The organisations highly differ in how they define checkworthiness characteristics and how they approach the choice of checkworthy claims. The authors of the TATHYA dataset conducted an empirical analysis of the dataset and found that fact-check organisations differ in the number and choice of claims that they fact-check \cite{patwari:17}. 
In the CW-USPD-2016, the authors also analysed annotation agreement between fact-check organisations and stated that agreement is low: only one out of 880 sentence is labeled as checkworthy by all nine sources, twelve sentences by seven sources and 97 sentences by four sources \cite{gencheva:17}. Fact-checkers seem to primarily focus on claims with questionable veracity, leading to initial veracity judgements and/or fact-check procedures that are not explicitly reported. This is particularly the case with FactCheck.org where factcheckers do not indicate which questioned, checkworthy claims appeared to have sufficient supporting evidence after contacting the person or organisation uttering the claim. Consequently, gold standard checkworthiness labels extracted from FactCheck.org (such as in CT-CWC-18 and CT19-T1) do not cover all checkworthy claims, but merely indicate checkworthy sentences that were not well and correctly corroborated by the speaker/writer afterwards. For example, the labeling choices of the following sequence of sentences, in which checkworthy sentences are in italics, call into question the difference between checkworthy and non-checkworthy, whereas some sentences are arguably comparable: 
\begin{quote}
    "I think the fact that -- that under this past administration was of which Hillary Clinton was a part, we've almost doubled the national debt is atrocious. I mean, I'm very proud of the fact that -- I come from a state that works. \textit{The state of Indiana has balanced budgets. We cut taxes, we've made record investments in education and in infrastructure, and I still finish my term with \$2 billion in the bank.} That's a little bit different than when Senator Kaine was governor here in Virginia. He actually -- he actually tried to raise taxes by about \$4 billion. He left his state about \$2 billion in the hole. \textit{In the state of Indiana, we've cut unemployment in half; unemployment doubled when he was governor.} " - Mike Pence, Vice-Presidential debate in the 2016 US campaign, labeled by FactCheck.org \cite{nakov:18}
\end{quote}
The checkworthiness of a claim also seems to entail temporarily popular topics and speakers. The popularity of a topic is characterized by its prevalence in the news at that moment and/or its demand and interest among the general public/readership. During the first labeling run of the ClaimBuster dataset, the authors observed that claims by more recent presidential candidates were more often labeled as checkworthy than earlier candidates \cite{hassan:17}, suggesting that checkworthiness indeed has a temporal dimension. \\ \\
In sum, fact-check organisations appear to differ in their approach to select checkworthy claims. Many base their selection on the perceived falsity of the conveyed information in a claim and on the popularity of topics in the news and/or public interest at the time of speaking/writing. As a result, checkworthiness has a strong temporal character and is dependent on how an individual perceives and estimates the veracity of a claim.
\subsection{Empirical Analysis of Checkworthiness} \label{graves}

In the previous section, we analysed the definitions of checkworthiness and checkworthy claim selection as mentioned on the website of several fact-check organisations, and briefly discussed how these are reflected in the datasets. In this section, we examine the characteristics of checkworthiness as defined by Graves \cite{graves:13} in the datasets. Graves~\cite{graves:13} observed political fact-checkers and elaborated on how fact-checkers at 'elite' fact-check organisations (FullFact.org, PolitiFact and The Washington Post's Fact Checker) choose which claims they are going to fact-check. He states that these organisations focus on objective claims uttered by political figures and organisations that have scientific validity and questionable veracity. We analyse if and how these characteristics are reflected in the datasets.

\begin{figure}%
    \centering
    \subfloat[checkworthy]{{\includegraphics[width=5.25cm]{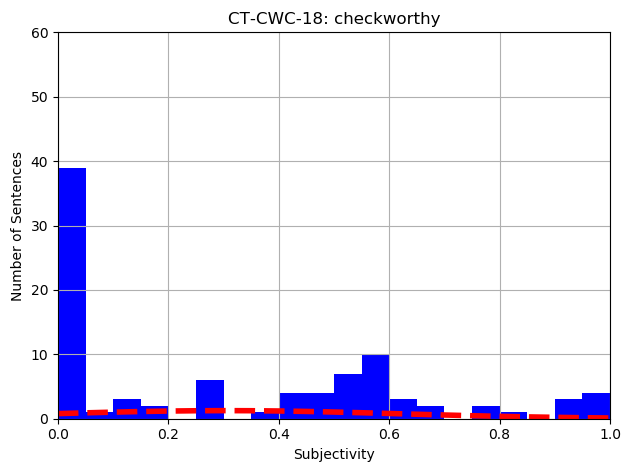} }}%
    \qquad
    \subfloat[non-checkworthy]{{\includegraphics[width=5.25cm]{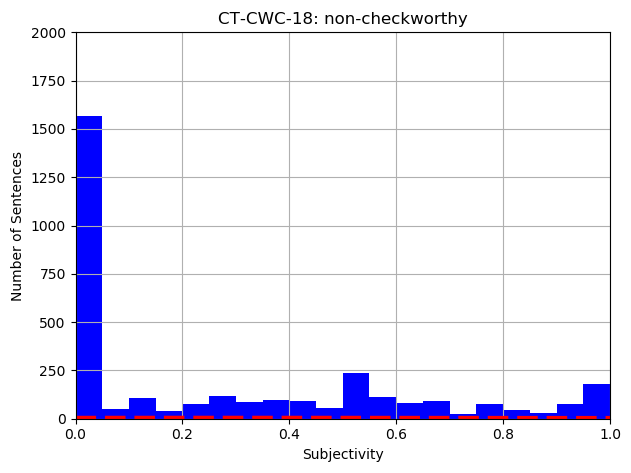} }}%
    \caption{Histogram of subjectivity scores of checkworthy (a) and non-checkworthy sentences (b) in CT-CWC-18}%
    \label{fig:subjectivity_clef}%
\end{figure}

\subsubsection{A checkworthy claim is always an objective statement} 
\begin{figure}%
    \centering
    \subfloat[checkworthy]{{\includegraphics[width=5.25cm]{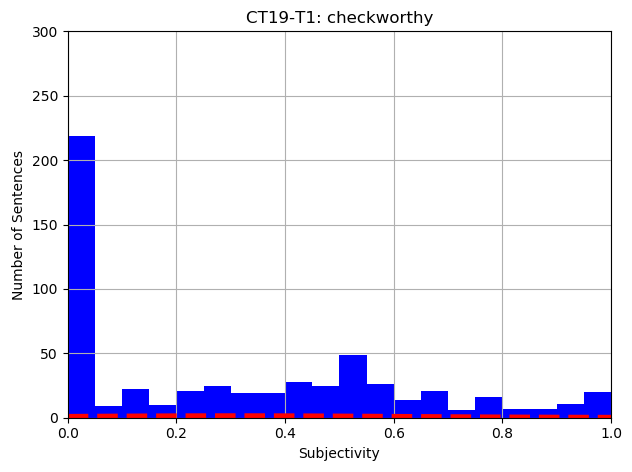} }}%
    \qquad
    \subfloat[non-checkworthy]{{\includegraphics[width=5.25cm]{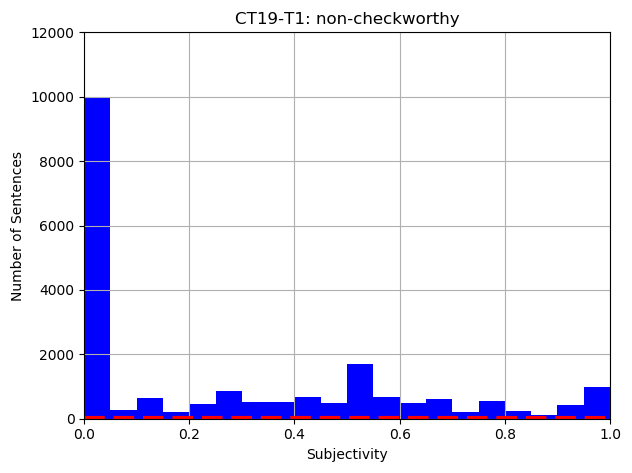} }}%
    \caption{Histogram of subjectivity scores of checkworthy (a) and non-checkworthy sentences (0) in CT19-T1}%
    \label{fig:subjectivity_clef_19}%
\end{figure}

and never a subjective statement such as opinions and speculations \cite{graves:13}. 
Hassan et al. \cite{hassan:17} ran a subjectivity classifier over a section of the ClaimBuster dataset. The results show that factual claims (UFS and CFS) are not exclusively classified as objective, and non-factual sentences (NFS) are more often classified as objective than subjective. They argue that a classification model basing it predictions solely on subjectivity and objectivity ratings cannot be used to separate non-factual sentences from factual, optionally checkworthy sentences. We examine whether similar observations can be made for the CT-CWC-18 and CT19-T1 datasets and run a subjectivity classifier over the two datasets. Each sentence in the dataset is assigned a subjectivity score using the sentiment library of the TextBlob package built on NLTK: a score of 0.0 denotes a highly objective sentence and a score of 1.0 a highly subjective sentence. Results are shown in Fig. \ref{fig:subjectivity_clef} and Fig. \ref{fig:subjectivity_clef_19}. About 45.65 \% and 41.99 \% of all checkworthy claims in, respectively, CT-CWC-18 and CT19-T1 have a subjectivity score of 0.1 or lower, while this is the case for 51.97 \% and 51.67 \% of all non-checkworthy claims in the respective datasets. It appears that the bulk of the sentences in both datasets is classified as objective and that there is no considerably large difference between the checkworthy and non-checkworthy class in terms of objectivity/subjectivity ratings. We, therefore, argue that most checkworthy claims are indeed objective, but that objectivity cannot be used as the only parameter to distinguish checkworthy from non-checkworthy claims. A number of checkworthy claims are labeled as (highly) subjective sentences. For example, the following two sentences, labeled as checkworthy, can be interpreted as speculations, whereas they talk about possible futures in which different tax plans would be implemented:
\begin{quote}
    \textit{"Independent experts have looked at what I've proposed and looked at what Donald's proposed, and basically they've said this, that if his tax plan, which would blow up the debt by \$5 trillion and would in some instances disadvantage middle-class families compared to the wealthy, were to go into effect, we would lose 3.5 million jobs and maybe have another recession. They've looked at my plans and they've said, OK, if we can do this, and I intend to get it done, we will have 10 million more new jobs, because we will be making investments where we can grow the economy."} - Hillary Clinton, First 2016 US Presidential Debate, labeled by FactCheck.org \cite{nakov:18}
\end{quote}
However, it can be checked whether independent experts indeed had made these statements.

\subsubsection{A checkworthy claim does not contain information that is common knowledge or self-evidently true} whereas these statements do not often entail false information \cite{graves:13}. This means that people need to make a first veracity judgement for each sentence in order to reject self-evidently true or commonly known claims for fact-checking. In the field of psychology, there has been ample research on how people judge veracity and how these judgements are influenced. If a statement is ambiguous and one is uncertain about a statement's truth value, one resides to heuristic cues using attributes such as source credibility, context in which the statement is presented, the statement itself and the metacognitive experience of fluency \cite{dechene:10}. Otherwise, the veracity judgement of a statement is mainly based on prior knowledge \cite{dechene:10}. However, prior knowledge does not necessarily constrain the influence of source credibility, context and fluency \cite{fazio:15}, especially when previous reliance on fluency has resulted in valid judgements \cite{scholl:14}. Types of fluency are processing fluency (i.e. the ease of processing), conceptual fluency (i.e. the ease of constructing a meaning and a more general semantic knowledge structure), perceptual fluency (i.e. the ease of perceiving) and linguistic fluency (i.e. the ease and simplicity of phonology) \cite{alter:09}. Fluency commonly results in a higher truth value \cite{fazio:15}. If we want to incorporate the ability of filtering commonly known or self-evidently true claims from factoid statements into a computational checkworthy claim detection model, the model needs to have access to the context in which the claim occurs and the credibility of the source. Even though the textual context in which the claims are uttered and the speakers are provided in most datasets, much more information about the context and the speakers is necessary to make a valid judgement, such as previous debates and political persuasion of the speakers. Concerning fluency, we doubt that a general, objective veracity judgement can be based on fluency, whereas the perception of fluency differs among individuals and is, therefore, subjective. Finally, if a model wants to judge the self-evident truth of a claim based on prior knowledge, it is basically performing a first fact-check procedure where it compares a claim to the veracity of previous, comparable claims and/or to a large database of, for example, texts. 
\begin{quote}
    "\textit{In El Paso, they have close to 2,000 murders right on the other side of the wall.} And they had 23 murders. \textit{It's a lot of murders, but it's not close to 2,000 murders right on the other side of the wall, in Mexico. So everyone knows that walls work.} And there are better examples than El Paso, frankly. You take a look. Almost everywhere. Take a look at Israel. They're building another wall. Their wall is 99.9 percent effective, they told me -- 99.9 percent." - Donald Trump, National Emergency Remarks (February 2019), labeled by FactCheck.org \cite{atanasova:19}
\end{quote}
The perceived veracity of the checkworthy claim \textit{"So everyone knows that walls work"} can thus depend on an individual's prior knowledge and judgement on walls around the world today and in the past, their veracity judgements of the murder numbers in El Paso and/or the effectiveness of the Israeli wall, their judgement on Donald Trump's credibility, the context in which Trump declares national emergency and fluency of the claim. \\ \\
During the annotation of the ClaimBuster dataset \cite{hassan:17,hassan:15}, the labeling choice of the participants was compared to that of three domain experts. Participants were then assigned a score between 0 and 1 according to their accordance with the domain experts. However, domain expertise does not necessarily mean that these experts have a more objective and correct first veracity judgments than non-experts. Moreover, it contradicts with the definition of checkworthiness given in the ClaimBuster papers, in which checkworthiness is the present appeal of the general public to recovering a claim's truthfulness. By ranking participants based on label agreement with expert domain labeling, the annotation may not reflect the general public's interest and appeal, but rather the interest and appeal of these three experts. As for the lower number of labeled checkworthy sentences in older transcripts in the ClaimBuster dataset, checkworthiness labeling may have been affected by both prior and posterior knowledge of the participants. In the CT-CWC-18 and CT19-T1 datasets, a checkworthiness label is automatically assigned to a sentence if it is fact-checked by FactCheck.org. An important shortcoming to this method has been discussed earlier in section \ref{factcheckorgan}: the fact-checkers from FactCheck.org do not report their first fact-check procedure, during which they fact-check statements that may be inaccurate by contacting the person/organisation from which the statements originates. In view of veracity judgements, this means that a number of claims with initially questioned veracity are wrongly contained within the non-checkworthy class. As a result, the non-checkworthy class consists of subjective, non-factual claims and objective statements that could be common knowledge or not.

\begin{figure}%
    \centering
    \subfloat[CT-CWC-18: non-checkworthy (11.2\% - 88.8\%)]{{\includegraphics[width=2.5cm]{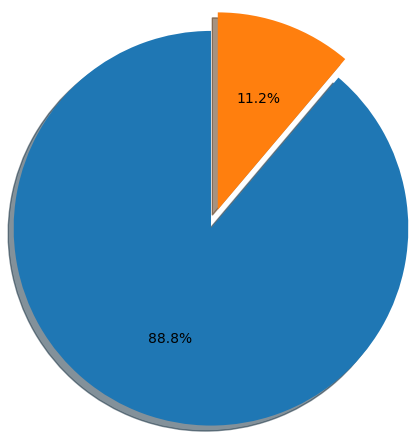}}}%
    \qquad
    \subfloat[CT-CWC-18: checkworthy (32.6\% - 67.4\%)]{{\includegraphics[width=2.5cm]{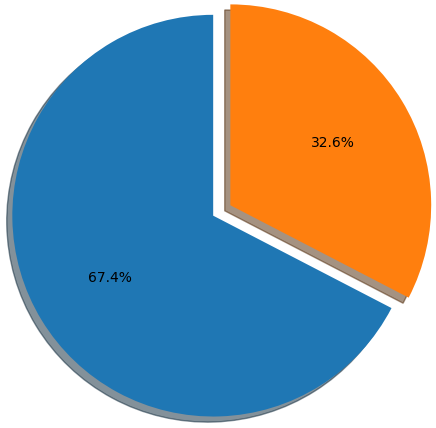}}}%
    \qquad
    \subfloat[CT19-T1: non-checkworthy (10.6\% - 89.4\%)]{{\includegraphics[width=2.5cm]{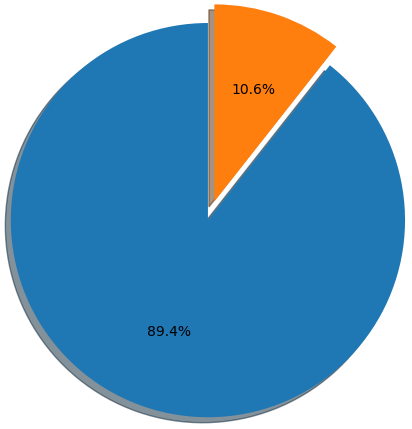}}}%
    \qquad
    \subfloat[CT19-T1: checkworthy (34\% - 66\%)]{{\includegraphics[width=2.5cm]{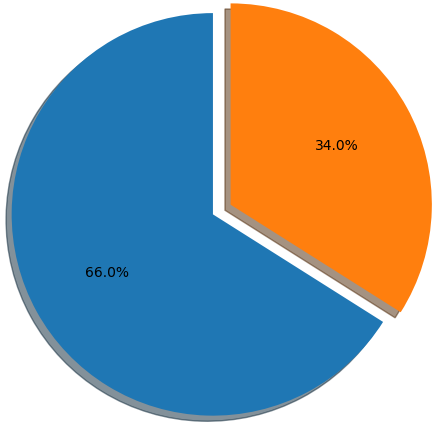}}}%
    \caption{Real numbers in the sentences of the CT-CWC-18 (a,b) and CT19-T1 (c,d) dataset (real numbers - no real numbers)}
    \label{fig:real_number}%
\end{figure}

\subsubsection{A checkworthy claim states a fact that has scientific validity} The claim reflects well the reality it supposedly represents and it has supporting scientific evidence for that representation \cite{graves:13}. In other words, the claim is 'checkable' and a veracity verdict is based on public data sources and independent experts \cite{graves:13}. For example, a claim containing real numbers can be checked against objective data. Real numbers can refer to, for example, a number of people, a percentage of income or a specific date. However, these numbers are subject to change and do not necessarily remain the same throughout time and space. For the CT-CWC-18 and CT19-T1 datasets, we compared the presence of real numbers - both written in numbers and words - in the checkworthy and non-checkworthy class (Fig. \ref{fig:real_number}). In CT-CWC-18, approximately one out of ten non-checkworthy sentences contains a cardinal number, with 1.4 numbers per sentence on average, while nearly one out of three checkworthy sentences have at least one cardinal number, with 1.9 numbers per sentence on average. Similar results are found in CT19-T1. Those results may indicate that checkworthy claims contain cardinal numbers more often than non-checkworthy claims and that the average number of cardinals is higher. Other checkable information such as place names, personal names and time indications might be more prominent in checkworthy sentences than in non-checkworthy sentences, but we leave this to further research.

\subsubsection{Professional fact-checkers and journalists pay attention to the individual or group/organisation uttering the statement} Especially statements from political figures are verified, while political news and statements of political persuasion are prone to deceiving and false information in order to persuade citizens of their convictions \cite{graves:13}. Apart from politicians, statements from labor unions and trade associations should be commonly fact-checked as well \cite{graves:13}. In this paper, however, we cannot check whether fact-checkers pay more attention to political figures and organisations, whereas the datasets predominantly consist of political debate transcripts where almost all claims are made by political figures. 
\section{Scope and Language Use} \label{scope_language}

The bulk of datasets for checkworthy claim detection consists of written transcripts of debates, speeches and/or conferences. Whereas oral language in the debates inherently differ from written language in news articles in their level of reciprocity (reciprocity between speakers and listeners vs. limited reciprocity between author and reader), discourse (primary vs. secondary), intersentence relations (paratactic vs. hypotactic) and cohesion cues (paralinguistic vs. lexical) \cite{horowitz:87}, it may not be straight-forward to transfer the models trained on transcripts of spoken language to predict checkworthiness in written language. Furthermore, the general aim of political debates and speeches to persuade the audience affects rhetorics and linguistics: speakers use more nominalization, passivization, metaphores, modality, parallellism (reiteration of similar syntactical and lexical units) and unification strategies (use of 'we' and 'our') \cite{kazemian:14}. The following excerpt, in which all sentences are labeled as checkworthy, is an example of parallellism:
\begin{quote}
    \textit{"But here, there was nothing to investigate from at least one standpoint. They didn't know the location. They didn't know the time. They didn't know the year. They didn't know anything."} - Donald Trump, UN press conference (September 2018), labeled by FactCheck.org \cite{atanasova:19}
\end{quote}
Several researchers constructing checkworthy claim detection models touch on the linguistic and rhetorical characteristics of debates and speeches, as well. Zuo, Karakas and Banerjee \cite{zuo:18}, who trained and tested their model on the CT-CWC-18 dataset, argue that the rhetorical and conversational features of debates, such as schesis onomation (= the repetition of synonymous expressions - predominantly nouns and adjectives - to emphasize and reinforce ideas), cause issues for the classification of checkworthy sentences. They also claim that the checkworthiness of many short sentences strongly depends on prior sentences and mention that interruptions by interlocutors or surroundings cause ill-formed or partly-formed sentences and discontinuous trains of thought. According to Hansen et al. \cite{hansen:19}, the higher number of raised topics in debates and the unbalanced representation of speakers cause their full neural checkworthiness model to perform worse on debates (0.0538 MAP) than on speeches (0.2502 MAP) in the CT19-T1 datasets. Not only the linguistic and rhetorical dynamics of debates and speeches, but also unbalanced speaker representation, low number of checkworthy claims and noisy data might explain those low MAP results. The next section elaborates on the latter, potential causes. All things considered, there need to be datasets containing diverse types of written and (transcribed) spoken language from different domains if a computational model is to detect checkworthy claims in various settings such as debates, speeches, news articles and blog posts. 

\section{Imbalance and Noise}

\begin{figure}
    \centering
    \includegraphics[width=10cm]{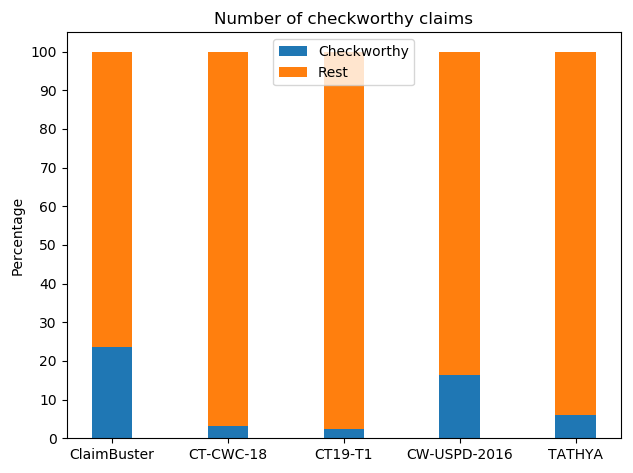}
    \caption{Imbalance}
    \label{fig:imbalance_sentences}
\end{figure}

The datasets are highly imbalanced in terms of checkworthy sentences, non-checkworthy sentences and speaker representation. Fig. \ref{fig:imbalance_sentences} displays the balance between checkworthy and non-checkworthy sentences in the five datasets. Overall, the majority of sentences is non-checkworthy and only a small share of the sentences is checkworthy, especially in the CT-CWC-18, CT19-T1 and TATHYA datasets. Consequently, classification models may have too few observations of the checkworthy class in order to accurately and precisely learn checkworthiness characteristics. Not only is there imbalance between the checkworthy and non-checkworthy class, but also the non-checkworthy class is highly noisy. This noise has several causes. Firstly, there is low label agreement due to diverging checkworthiness definitions. Secondly, a notable number of checkworthy claims are present in the non-checkworthy class whereas these are considered commonly know and/or self-evidently true or appear to be truthful after corroboration by the speaker - the latter is the case in the CT-CWC-18 and CT19-T1 datasets. Although a claim's checkworthiness is dependent on prior and, possibly, posterior sentences, annotators are asked to judge and label the checkworthiness of a claim of which the context is omitted by default. That is, especially, the case in the ClaimBuster dataset. Thirdly, labeling entire sentences as checkworthy leads to inclusion of non-checkworthy information in the checkworthy class.

{}

As for speaker representation, imbalance in the dataset can affect bias towards or against speakers and, thus, overall model performance. Speakers may differ in lexicon, syntax, rhetorics, points-of-view and topics of interest. Table \ref{tab:distribution_18} and \ref{tab:distribution_19} display the speaker distributions in CT-CWC-18 and CT19-T1, respectively. The audience and moderators are left out, whereas the fact-checkers at FactCheck.org mainly focused on the politicians participating in the debates. It appears that Trump is strongly represented in both datasets: 34.52 \% and 44.10 \% of all sentences in, respectively, CT-CWC-18 and CT19-T1 are uttered by Trump. This is due to the presence of his speeches in both datasets. It might be beneficial for model performance to include speeches by other politicians, especially from those representing other political stances.

\begin{table}[]
    \centering
    \begin{tabular}{|c|c|c|c|c|c|}
        \hline
        Speaker & \# Sentences & Share in Dataset & \# Checkworthy & Share in Dataset & CW/tot. sent. \\
        \hline
        \hline
        Trump & 1403 & 34.52 \%  & 49 & 52.13 \% & 3.49 \% \\
        Clinton & 790 & 19.46 \%  & 15 & 15.96 \% & 1.90 \% \\
        Kaine & 600 & 14.76 \%  & 19 & 20.21 \% & 3.17 \% \\
        Pence & 524 & 12.89 \%  & 9 & 9.57 \% & 1.72 \% \\
        \hline
    \end{tabular}
    \caption{Speaker respresentation in CT-CWC-18}
    \label{tab:distribution_18}
\end{table}

\begin{table}[]
    \begin{tabular}{|c|c|c|c|c|c|}
        \hline
        Speaker & \# Sentences & Share in Dataset & \# Checkworthy & Share in Dataset & CW/tot. sent. \\
        \hline
        \hline
        Trump & 10363 & 44.10 \%  & 398 & 69.10 \% & 3.84 \% \\
        Clinton & 2485 & 10.57 \%  & 60 & 10.42 \% & 2.41 \% \\
        Sanders & 1202 & 5.11 \%  & 23 & 3.99 \% & 1.91 \% \\
        Rubio & 823 & 3.50 \%  & 22 & 3.82 \% & 2.67 \% \\
        Cruz & 690 & 2.94 \% & 21 & 3.65 \% & 3.04 \% \\
        Kasich & 643 & 2.74 \% & 4 & 0.69 \% & 0.62 \% \\
        Kaine & 600 & 2.55 \% & 19 & 3.30 \% & 3.17 \% \\
        Pence & 524 & 2.23 \% & 9 & 1.56 \% & 1.72 \% \\
        O'Malley & 230 & 0.98 \% & 5 & 0.87 & 2.17 \% \\
        Carson & 177 & 0.75 \% & 0 & 0 \% & 0 \% \\
        Bush & 144 & 0.61 \% & 0 & 0 \% & 0 \% \\
        Christie & 127 & 0.54 \% & 0 & 0 \% & 0 \% \\
        Paul & 116 & 0.49 \% & 0 & 0 \% & 0 \% \\
        Pelosi & 97 & 0.41 \% & 1 & 0.17 \% & 1.03 \% \\
        Schumer & 66 & 0.28 \% & 1 & 0.17 \% & 1.52 \% \\
        \hline
    \end{tabular}
    \caption{Speaker representation in CT19-T1}
    \label{tab:distribution_19}
\end{table}
\section{Conclusion}

It can be argued that there are several shortcomings in automated checkworthy claim detection research. Firstly, there is insufficient agreement on the definition and characteristics of checkworthiness and how a checkworthy claim can be differentiated from a non-checkworthy claim. Drawing from dictionary entries on \textit{check}, \textit{worthiness} and \textit{worthy}, we provided a preliminary, rather abstract delineation of checkworthiness, where we defined it as a the concept of having a certain worth, value, suitability and need to have the correctness of what it portrays or entails verified. Given the definitions applied by several fact-check organisations in combination with computational journalism research and psychology, we specify checkworthiness as follows: checkworthiness is the concept of having a time-dependent, space-dependent and context-dependent worth, value, suitability, ability and need to have the correctness of the objectivity it conveys verified, irrespective of its perceived veracity judgement by an individual based on prior knowledge and beliefs. Aside from the diverging definitions of checkworthiness, we come to the conclusion that current datasets are too limited in scope and language, and that the classes are too noisy. Therefore, we do not only deem current datasets used for training and testing the models insufficient for the task, but we also argue that the checkworthiness detection task itself is too subjective and can thus not be objectively approached by computational models. However, it would be too computationally demanding to check every objective, factual claim in each text. We, therefore, suggest that it might be better to apply other filtering methods instead of a checkworthiness filter, such as speculation detection \cite{cruz:16,qian:16}, rumour detection \cite{zubiaga:18} and/or topic-dependent claim detection \cite{levy:14}.

\bibliographystyle{splncs04}
\bibliography{mybibliography}


\end{document}